\documentclass[11pt,a4paper]{article}

\usepackage{times,latexsym}
\usepackage{url}
\usepackage[T1]{fontenc}
\usepackage[utf8]{inputenc}
\usepackage{microtype}
\usepackage{inconsolata}
\usepackage{graphicx}

\usepackage{booktabs}
\usepackage{pgf}
\usepackage{glossaries-extra}
\usepackage{makecell}
\usepackage{float}
\usepackage{tikz}
\usepackage{musicography}
\usepackage{tcolorbox}
\usepackage{enumitem}
\usetikzlibrary{arrows.meta,patterns,patterns.meta}

\setabbreviationstyle[acronym]{long-short}
\glsdisablehyper

\newacronym[
  longplural={large language models},
  shortplural={LLMs}
]{llm}{LLM}{large language model}

\newacronym[
  longplural={multimodal large language models},
  shortplural={MLLMs}
]{mllm}{MLLM}{multimodal large language model}

\newacronym{ai}{AI}{artificial intelligence}

\newacronym[
  longplural={generalised linear mixed models},
  shortplural={GLMMs}
]{glmm}{GLMM}{generalised linear mixed model}

\newacronym{fpr}{FPR}{false positive rate}

\usepackage[acceptedWithA]{tacl2021v1}
\hypersetup{
  pdftitle={Psychophysics of Duration Perception in Multimodal Large Language Models},
  pdfauthor={Auss Abbood, Zihao Fu, Nigel Collier},
  pdfsubject={Preprint}
}

\title{Psychophysics of Duration Perception \\
in Multimodal Large Language Models}
\author{%
  Auss Abbood \\
  {\normalfont Language Technology Lab} \\
  {\normalfont University of Cambridge} \\
  {\normalfont\texttt{aa2613@cam.ac.uk}}
  \And
  Zihao Fu \\
  {\normalfont The Chinese University} \\
  {\normalfont of Hong Kong} \\
  {\normalfont\texttt{zihaofu@cuhk.edu.hk}}
  \And
  Nigel Collier \\
  {\normalfont Language Technology Lab} \\
  {\normalfont University of Cambridge} \\
  {\normalfont\texttt{nhc30@cam.ac.uk}}%
}
\date{}

\begin{document}
\maketitle
\thispagestyle{plain}
\begin{abstract}  
 Studies of time perception in language models focus mainly on text-based settings.
 Yet multimodal models are increasingly used for tasks that require perceiving the passage of time, making the limits of their temporal perception important to understand.
 We measure the limits of duration perception from audio against a human baseline to address this gap.
 Drawing from psychophysics, we operationalise duration perception as a melody recognition task, testing when a melody becomes unrecognisable as it is slowed down.
 Of 21 tested models, most cannot recognise well-known melodies without labelled in-context examples. With in-context examples, seven recognise melodies reliably, most peaking within the median human recognition band.
 Generalisation is limited: (i) recognition peaks shift with in-context example speed, (ii) melody titles induce prediction bias, and (iii) providing examples with scientific pitch notation does not prevent accuracy declining at long note durations for most models.
 In humans, the recognition window is a property of the perceiver; in current models, it is largely a property of the prompt.
\end{abstract}

\section{Introduction}

There is increasing interest in understanding time perception in \glspl{llm} \citep{chenPerceivePassageTime2025}. 
Existing approaches can be largely separated into two strands: 
One probes duration perception on the discourse level, measuring its alignment with humans' subjective judgements of duration or urgency \cite{wangDiscreteMindsContinuous2025, chengYourLLMAgents2026} (``\textit{What feels longer, A or B?}''). 
The other strand investigates \glspl{llm}['] ability to estimate task duration and calibration on the agentic level \citep{garikaparthiCANLLMSPERCEIVE2026, maksymandriushchenkoYourAgentsAre2026} (``\textit{How long do you need for this task?}'').
However, the limits of time perception in \glspl{mllm}---and how they compare with human limits---remain poorly understood.
We investigate these limits using a melody recognition task that combines psychophysics and Gestalt psychology \citep{gildenPsychologyHumanTemporality2025}.

Melody recognition survives temporal rescaling~\citep{lerdahlGenerativeTheoryTonal1996}: a familiar tune remains recognisable at various pitches and tempi~\citep{tenneyTemporalGestaltPerception1980}.
Our perception limit is demarcated by when a familiar tune \emph{stops} surviving temporal rescaling.

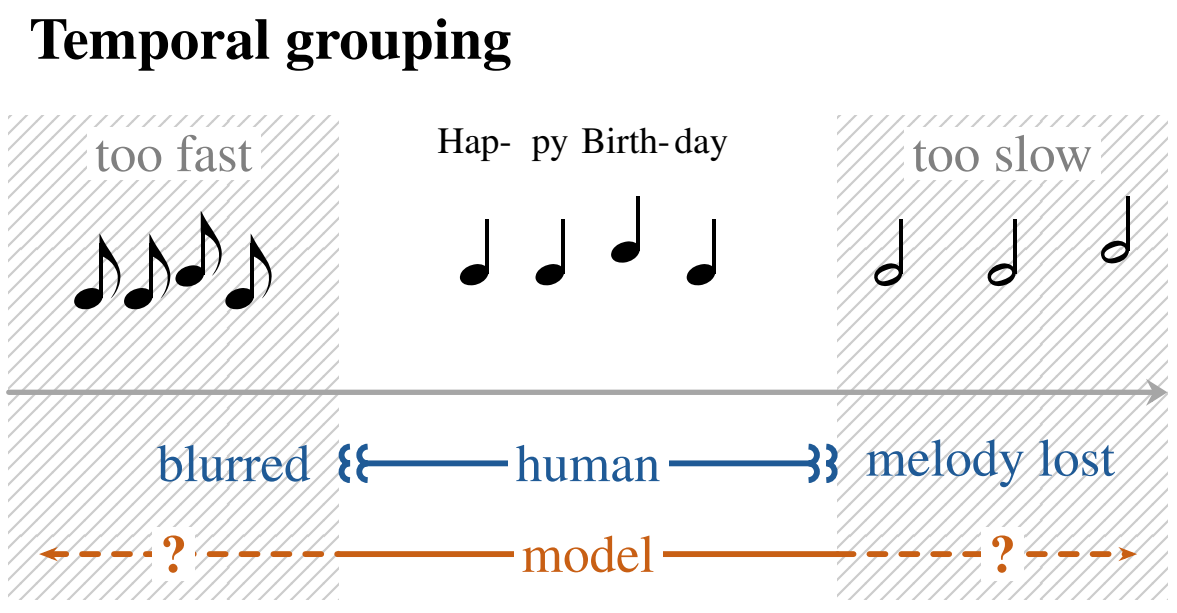
\begin{figure}[!t]
  \centering
  % TikZ/PGF fragment: Temporal Gestalt grouping limits.
% Required in the main document preamble:
%   \usepackage{tikz}
%   \usepackage{musicography}
%   \usetikzlibrary{arrows.meta,patterns,patterns.meta}

\begingroup%

\definecolor{humancol}{RGB}{31,90,150}%
\definecolor{modelcol}{RGB}{200,95,20}%

% Locally scoped helper commands with figure-specific names.
\newcommand{\gestaltenote}[3]{%
  \node[inner sep=0pt, anchor=center, scale=#3] at (#1,#2) {\musEighth};%
}%
\newcommand{\gestaltqnote}[3]{%
  \node[inner sep=0pt, anchor=center, scale=#3] at (#1,#2) {\musQuarter};%
}%
\newcommand{\gestalthnote}[3]{%
  \node[inner sep=0pt, anchor=center, scale=#3] at (#1,#2) {\musHalf};%
}%

% Layout knobs (cm).
% XMin/XMax extend the original 0..7.1 range by 0.282 cm on each side, which
% makes the drawing (including the label overhang) exactly 7.7 cm wide so it
% fills the \makebox below edge to edge. Extending both ends symmetrically
% keeps the hatched zones equal in width and leaves the midpoint (and thus
% the centred "human"/"model" labels and all drawn elements) unchanged.
\newcommand{\GestaltXMin}{-0.282}%
\newcommand{\GestaltLo}{1.9}%
\newcommand{\GestaltHi}{5.2}%
\newcommand{\GestaltXMax}{7.382}%
\pgfmathsetmacro{\GestaltMid}{(\GestaltXMin+\GestaltXMax)/2}% axis / label centre
\newcommand{\GestaltHumanY}{-0.15}% blue human range bar
\newcommand{\GestaltModelY}{-0.75}% orange model range bar
\newcommand{\GestaltZoneBottom}{-1.05}% hatching extends below both range rows

% The outer box is exactly 7.7 cm wide. Nothing inside is scaled, so font
% sizes remain true to the values specified below.
\makebox[7.7cm][c]{%
\begin{tikzpicture}[
  font=\fontsize{10}{12}\selectfont,
  line cap=round
]

\tikzset{
  bar/.style={line width=1.1pt},
  endtick/.style={line width=1.1pt},
  axis/.style={
    ->,
    >={Stealth[length=4pt, width=5pt]},
    gray!70,
    line width=1pt
  },
  zone/.style={
    pattern=north east lines,
    pattern color=gray!40,
    draw=none
  },
  lbl/.style={inner sep=1.5pt, fill=white},
  zlbl/.style={gray, align=center, fill=white, inner sep=1.5pt},
  title/.style={
    anchor=west,
    font=\fontsize{11}{13}\selectfont\bfseries
  },
  rangecap/.tip={Hooks[sep] . Hooks[]},
}

% Temporal grouping
\begin{scope}
  \node[title] at (\GestaltXMin,2.62) {Temporal grouping};

  \fill[zone] (\GestaltXMin,\GestaltZoneBottom) rectangle (\GestaltLo,2.15);
  \node[zlbl] at ({(\GestaltXMin+\GestaltLo)/2},1.90) {too fast};

  \fill[zone] (\GestaltHi,\GestaltZoneBottom) rectangle (\GestaltXMax,2.15);
  \node[zlbl] at ({(\GestaltHi+\GestaltXMax)/2},1.90) {too slow};

  \foreach \x/\syl in {2.8/Hap-, 3.3/py, 3.8/Birth-, 4.3/day} {%
    \node[
      anchor=base,
      inner sep=1pt,
      font=\fontsize{7}{6.5}\selectfont
    ] at (\x,1.9) {\syl};%
  }

  % Left/right note clusters are centred in the hatched zones; after the
  % ±0.282 cm frame extension that means a ±0.141 cm shift from the old
  % 0..7.1 coordinates. Middle (quarter) notes sit between Lo/Hi and are unchanged.
  \foreach \x/\y in {0.309/1.1, 0.639/1.1, 0.979/1.25, 1.309/1.1}
    {\gestaltenote{\x}{\y}{1.6}}
  \foreach \x/\y in {2.8/1.1, 3.3/1.1, 3.8/1.25, 4.3/1.1}
    {\gestaltqnote{\x}{\y}{1.6}}
  \foreach \x/\y in {5.541/1.1, 6.291/1.1, 7.041/1.25}
    {\gestalthnote{\x}{\y}{1.6}}

  \draw[axis] (\GestaltXMin,0.32) -- (\GestaltXMax,0.32);
%  \node[gray, inner sep=1.5pt] at (\GestaltMid,0.53)
%    {note duration};
%  \draw[gray!70, line width=0.5pt] (\GestaltLo,0.27) -- (\GestaltLo,0.37);
%  \node[gray, anchor=north] at (\GestaltLo,0.25) {160\,ms};
%  \draw[gray!70, line width=0.5pt] (\GestaltHi,0.27) -- (\GestaltHi,0.37);
%  \node[gray, anchor=north] at (\GestaltHi,0.25) {2\,s};

  \draw[bar, humancol, rangecap-rangecap]
    (\GestaltLo,\GestaltHumanY) -- (\GestaltHi,\GestaltHumanY);
  % \draw[endtick, humancol] (\GestaltLo,-0.45) -- (\GestaltLo,-0.25);
  % \draw[endtick, humancol] (\GestaltHi,-0.45) -- (\GestaltHi,-0.25);
  \node[text=humancol, anchor=east] at ({\GestaltLo-0.05},\GestaltHumanY) {blurred};
  \node[lbl, text=humancol] at (\GestaltMid,\GestaltHumanY) {human};
  \node[humancol, anchor=west] at ({\GestaltHi+0.05},\GestaltHumanY) {melody lost};

  \draw[bar, modelcol] (\GestaltLo,\GestaltModelY) -- (\GestaltHi,\GestaltModelY);
  \draw[
    bar,
    modelcol,
    dashed,
    ->,
    >={Stealth[length=3.5pt]}
  ] (\GestaltHi,\GestaltModelY) -- ({\GestaltXMax-0.2},\GestaltModelY);
  % \draw[endtick, modelcol] (\GestaltLo,-1.05) -- (\GestaltLo,-0.85);
  % \node[text=modelcol, anchor=east] at ({\GestaltLo-0.05},\GestaltModelY) {tokeniser lim.};
  \draw[
    bar,
    modelcol,
    dashed,
    <-,
    >={Stealth[length=3.5pt]}
  ] ({\GestaltXMin+0.2},\GestaltModelY) -- (\GestaltLo,\GestaltModelY);
  \node[lbl, text=modelcol] at (\GestaltMid,\GestaltModelY) {model};
  \node[lbl, text=modelcol] at ({(\GestaltXMin+\GestaltLo)/2},\GestaltModelY) {\textbf{?}};
  \node[lbl, text=modelcol] at ({(\GestaltHi+\GestaltXMax)/2},\GestaltModelY) {\textbf{?}};

\end{scope}

\end{tikzpicture}%
}%

\endgroup
  \caption{Conceptual illustration of temporal Gestalt grouping
    limits in humans and models. Blue and orange indicate human and model
    ranges, and hatched regions fall outside the human grouping range. Hooked
    endpoints denote approximate boundaries. Question marks indicate unknown model limits.}
  \label{fig:gestalt-limits}
\end{figure}

Figure~\ref{fig:gestalt-limits} illustrates temporal grouping in melody recognition and contrasts human grouping limits with the unknown limits in models.
At very short note durations, melody recognition fails due to physiological limitations~\citep{hsiehTemporalIntegrationAbsolute2007, alves-pintoForwardMaskingEstimated2010, urataniSuppressionSubsequentN1m2014}.
However, above 1--2\,s, melody recognition fails even though every note is perfectly audible \citep{warrenMelodicNonmelodicSequences1991, andrewsIdentificationSpeededSlowed1998, dowlingMelodyRecognitionFast2008}; what dissolves is the tie between them (Fig.~\ref{fig:gestalt-limits}).
This boundary defines our duration perception limits.

A transformer-based model has no obvious reason to have a perceptual ceiling similar to that of humans.
It attends to the whole melody at once; recognising a slowed-down or sped-up melody is symbolically trivial.
To this end, we assess 21 audio and omni models using the eight melodies of \citet{warrenMelodicNonmelodicSequences1991} and verify our results on an expanded set of eleven more well-known melodies.
None of the tested models reliably recognise the melodies from parametric knowledge alone at normal speed, so we could not use the same setup as the human study.
We therefore provide labelled candidate melodies in context.
Seven models qualify for the main experiment.

For most qualifying models, recognition initially peaks at note durations similar to those observed in humans.
However, varying the speed of the in-context examples shifts this peak, with reliable recognition confined to a narrow range around the examples' note durations.
Melody labels also affect recognition.
For example, replacing melody titles with neutral codes reduces prediction bias and improves accuracy.
Together, these findings suggest that the apparent temporal perception limits depend strongly on the context provided, making it difficult to identify the models’ intrinsic perceptual limits.

Our contributions can be summarised as follows:

\begin{enumerate}[parsep=0pt,partopsep=0pt,leftmargin=*]
  \item We introduce a \textbf{psychophysical approach to measuring temporal perception limits in \glspl{mllm}}, using melody recognition to examine how duration affects recognition relative to a human baseline.
  \item We show that \textbf{apparent temporal recognition limits shift with in-context example speed}, challenging their interpretation as stable perceptual boundaries.
  \item We show that \textbf{melody labels can bias auditory recognition}, with neutral codes reducing prediction bias and improving accuracy.
\end{enumerate}

\section{Background}

\paragraph{Time perception in LLMs}

Research on temporal reasoning in \glspl{llm} has overlooked time perception \citep{chenPerceivePassageTime2025}.
Subsequent work filled this gap by operationalising time perception as understanding temporal relativity for example in context-dependent judgements tasks \citep{chenPerceivePassageTime2025}, or judging elapsed time in narrative settings \citep{baoMachinesInternalClock2026}.
Limits in time perception are attributed to the need of mapping continuous time to discrete tokens \citep{wangDiscreteMindsContinuous2025}.

Time perception limits keep resurfacing in agentic settings.
Limits manifest as the inability to align tools calls with human time sensitivity \citep{chengYourLLMAgents2026} or overestimating wall-clock time for agentic tasks by 4--7 times
\citep{garikaparthiCANLLMSPERCEIVE2026}.

All studies find limitations in \glspl{llm} and agents to ground time perception through token alone.
A possible explanation is that time perception requires multimodality or world models \citep{garikaparthiCANLLMSPERCEIVE2026}.
Therefore, we focus on \glspl{mllm}.

\paragraph{Auditory psychophysics in LLMs}

Early psychophysical test on \glspl{llm} found merit in this method: they found that human perception properties can be recovered from \glspl{llm} trained on text alone \citep{marjiehLargeLanguageModels2024}.

As \glspl{mllm} become more established, perception tests are increasingly applied to multimodal data. 
In the auditory domain, SonicBench uses controlled stimuli to test absolute and comparative judgements of physical properties, including duration and tempo \citep{sunSonicBenchDissectingPhysical2026}. 
Its stimuli are designed to be well above human discrimination thresholds, so it tests coarse judgements of individual acoustic properties rather than estimating model thresholds or limits of time perception.

Music-related benchmarks have overlap with psychophysical experiments in that they also test perception limits in \glspl{mllm}.
\citet{caroneLLMsCanRead}, for example, test models on stimuli with varying rhythm, melody, and harmony.
They find that models generalise poorly to transposed melodies presented as audio but perform near ceiling on MIDI (a symbolic music representation), concluding that models struggle to \emph{listen} to music.
The MUSE Benchmark test relational reasoning in music \citep{caroneMuseBenchmarkProbing2026}.
Among its tasks, it tests the recognition of grouping of beats into threes, fours, or fives.
However, these are categorical judgements on a fixed stimulus set; MUSE does not vary note duration to estimate when temporal grouping breaks down.

Existing work therefore tests either physical duration judgements or musical grouping, but not the psychophysical limits of duration-dependent temporal grouping. 
Our work fills this gap.

\paragraph{Gestalt in LLMs}

Time perception in humans is commonly explained via internal-clock theories \citep{vanrijnDedicatedClockTimingcircuit2014}, such as Scalar Timing Theory, which posits the existence of an internal pacemaker whose pulses anchor the flow of time.
\citet{gildenPsychologyHumanTemporality2025} contrasts this theory with a Gestalt account in which temporal grouping explains phenomena not captured by internal-clock theory, such as the loss of temporal grouping shown in Fig.~\ref{fig:gestalt-limits}.
Existing work on Gestalt and \glspl{mllm} has focused primarily on vision \citep{houSeeingBelievingBenchmark2026, shaGestaltVisionDataset}.
Our study is the first to use a Gestalt account to test duration perception in \glspl{mllm}.

\section{Duration perception probe}
\label{sec:duration-probe}

Our experimental setup is derived from the study by \citet{warrenMelodicNonmelodicSequences1991}. 
In short, it investigated the limits of temporal grouping in human temporal perception by letting humans classify well-known melodies at varying speeds.
The point at which notes stop grouping into a melodic Gestalt and are instead perceived individually \cite{tenneyTemporalGestaltPerception1980} denotes the perception limit. 
In humans, the median temporal perception window lies between 160\,\textup{ms} and 1280\,\textup{ms} (first quartile at 113\,\textup{ms} and third quartile at 2560\,\textup{ms}). 
The loss of temporal grouping at increased note durations, unlike the physiological limits at
short ones, indicates a limit in our temporal perception.

The experiment is divided into three steps: training, screening and duration scaling. For each step, we describe our setup and compare it with the original human study.

\paragraph{Melodies}
We use the same eight melodies as the original study by \citet{warrenMelodicNonmelodicSequences1991}: Twinkle, Twinkle, Little Star, Happy Birthday, God Rest Ye Merry Gentlemen, Rock-a-bye Baby, Camptown Races, Yankee Doodle, Love Me Tender, and Skip to My Lou.

Each melody \(\ell_i\) is represented as an ordered sequence of pitches,
\[
  \ell_i = (p_{i,1}, p_{i,2}, \dots, p_{i,n_i}),
  \quad n_i \in \{6,7,8,9\}.
\]
Melodies are presented isochronously, i.e., all notes in a stimulus have the same duration \(d_j\), where \(d_j \in D\). The set of possible note durations is
\begin{equation}
  \begin{aligned}
    D ={} &
    \{\operatorname{round}(40(\sqrt{2})^k)\,\mathrm{ms}
    : k \in \{0,\dots,12\}\}              \\
          & {}\cup \{3600\,\mathrm{ms}\}.
  \end{aligned}
  \label{eq:duration-set}
\end{equation}
The final stimulus presented to the model is defined by its melody and duration,
\[
  s_{ij} = (\ell_i,d_j).
\]

The melodies used, their scientific pitch notation, and an extended list of tested melodies are given in Appendix~\ref{app:melodies}. To avoid recognition by the first notes alone, the original study and this one use different entry points into a melody and then cyclically shift the sequence. For example, starting at the third pitch gives
\[
  (p_{i,3}, p_{i,4}, \dots, p_{i,n_i}, p_{i,1}, p_{i,2}).
\]
A melody may be looped, playing the same sequence of pitches \(n\) times. With a single loop, the sequence is played twice.

Melodies were rendered as sine waves. The exact setup is described in Appendix~\ref{app:melodies}.

\paragraph{Training} This stage ensures that the tested subject can recognise the melody before any temporal dilation or contraction is applied. In the human study, melodies were played at a \emph{normal} speed of 320\,\textup{ms} per note. Unless stated otherwise, normal speed refers to this value. Human participants needed to classify melodies correctly with the help of a list that contained the names of the tested melodies. Only if they recognised the melodies did they continue to the next stage.

We test two variations of the training phase for \gls{ai} models. The first variant presents melodies in a \emph{zero-shot}, multiple-choice format, testing whether a model can recognise the melody from its parametric knowledge alone. The second variant frames the training phase as \emph{in-context learning}, where we demonstrate the target and distractor melodies in context before asking the model to classify the target melody. This setting may bridge shortcomings in the model's ability to recognise a melody from its parametric knowledge alone.

\paragraph{Screening} In the human study, participants qualified for the main experiment if they could classify melodies without the aid of a melody list. During screening, melodies were again played at normal speed, but now starting at the third or fourth note to rule out recognition by the first note alone. The melody was looped without pauses until the human subject made a classification. Subjects were included in the main study if they recognised six out of eight melodies.

We adopt this design, with an adaptation of the looping setup. An interactive classification as described in the human study is uncommon and difficult to implement in practice. Instead, we apply a static variation in which the target melody is looped once --- the minimum used in the original study. Limiting presentation to one loop also ensures comparability and caps test duration and costs. Reference melodies start at the first note and are played once. Each candidate set therefore contributes 16 trials (8 melodies $\times$ entry at note 3 or 4), and we scale the human acceptance criterion of 6/8 correct classifications to 12/16 trials.

Note that in the in-context version of the training phase, the training and screening phases are done in one go: providing in-context examples is the training phase and requesting models to make a prediction based on the looped melody starting from the third or fourth note is the screening phase.

Lastly, we test two formats for the screening experiment: \emph{audio-only} and \emph{text+audio}. The text+audio format uses a mix of text prompts with interleaved audio files for the reference and target melodies. Since not all models accept this input format, either because they accept audio input only, or because they are limited by the number of audio files they can process in a single prompt, we also translate all experiments into an audio-only format. Interleaved audio-text inputs can be challenging because audio produces long token sequences, and models must preserve the relationships between alternating textual labels and audio clips \citep{zhaoATIRAudioTextInterleaved2026}. We use text-to-speech to translate the instructions to audio with minor adaptations of the instructions to fit the spoken setting. The audio-only setting produces a single audio file with spoken instructions and played melodies. Exact prompts and further details are in Appendix~\ref{app:prompts}.

\paragraph{Duration scaling} In the human study, half the participants first heard the target melody at 40\,\textup{ms} per note. Note duration was then multiplied by $\sqrt{2}$ until the melody was recognised. The other half started at 3600\,\textup{ms} and note duration was divided by $\sqrt{2}$ until recognition.

During duration scaling, human participants heard each melody in a continuous loop until they made a classification. Model trials are independent, so we instead test every note duration in Eq.~\ref{eq:duration-set} and use a fixed presentation: each target melody starts at either the third or fourth note and is looped once. This gives two trials at each of the 14 note durations, or 28 trials per melody.

Musicians are able to solve this task with much higher accuracy, even at long note durations \citep{warrenMelodicNonmelodicSequences1991}, because they can transcribe the melody as they listen to the recording, compensating for any perception limitations. We test whether \glspl{llm} are equally capable of overcoming such limitations with the help of transcription. In a transcription variant, we provide each candidate melody's scientific pitch notation (see Tabs.~\ref{tab:melodies-original} \& \ref{tab:melodies-extended}).

\paragraph{Evaluation} Our experiments are designed as multiple-choice, asking the model to classify the target melody among the options given in context and to respond with that melody's name only. Because some melodies go by other names, we recognise alias names for melodies, too. For example, Twinkle, Twinkle, Little Star has the same melody as the ABC Song. We report the alias-corrected accuracy scores (more details in Appendix~\ref{app:scoring}). For the screening experiments, we test different numbers of in-context examples. To maintain comparability, we report chance-corrected accuracy in percent, defined as
\[
  100 \times \frac{\text{accuracy} - \text{chance}}{1 - \text{chance}},
\]
where accuracy is the observed proportion correct and chance is the guessing probability determined by the number of in-context melodies. The chance-corrected 12/16 thresholds are 66.7\%, 70.0\%, and 71.4\% for 4, 6, and 8 candidates, respectively.

We further analyse the detection envelope of \glspl{llm} to characterise how accuracy in melody-recognition changes with note duration. To this end, we model trial-level correctness using a Bernoulli \gls{glmm} with a logit link. For stimulus \(s\) and model \(m\), let \(y_{s,m} \in \{0,1\}\) indicate task success, and let \(d_s\) denote the note duration in seconds. We fit
\begin{equation}
  \begin{aligned}
    y_{s,m} \mid p_{s,m} & \sim \operatorname{Bernoulli}(p_{s,m}), \\
    \operatorname{logit}(p_{s,m})
                         & =
    \beta_0
    +
    f\!\left(\log d_s\right)
    +
    u_m
    +
    v_{\ell[s]}.
  \end{aligned}
  \label{eq:duration-glmm}
\end{equation}
Here \(f(\cdot)\) is a cubic B-spline basis expansion with \(K\) degrees of freedom, \(u_m\) is a random intercept for model identity, and \(v_{\ell[s]}\) is a random intercept for melody identity. The random intercepts are assumed to follow
\[
  u_m \sim \mathcal{N}(0,\sigma^2_{m}),
  \qquad
  v_{\ell} \sim \mathcal{N}(0,\sigma^2_{\ell}).
\]
We choose the spline degrees of freedom as described in Appendix~\ref{app:glmm}.

\paragraph{Models} We test a variety of closed- and open-source models, some of which are multimodal or audio only. These include Gemini 3.1 Pro Preview~\cite{googleGemini31ProModelCard2026}, Gemini 3 Flash Preview~\cite{googleGemini3FlashModelCard2025}, Gemini 2.5 Pro and 2.5 Flash~\cite{geminiTeamGemini25TechnicalReport2025}, and Gemini 2.5 Flash-Lite~\cite{googleGemini25FlashLiteModelCard2025}; GPT Audio~\cite{openaiGPTAudio} and GPT Audio Mini~\cite{openaiGPTAudioMini}; Qwen3.5 Omni Plus, 3.5 Omni Flash~\cite{qwenteamQwen35OmniTechnicalReport2026}, 3 Omni Flash, 3 Omni 30B A3B Thinking and Instruct~\cite{xuQwen3OmniTechnicalReport2025}, 2 Audio 7B Instruct~\cite{chuQwen2AudioTechnicalReport2024}; Voxtral Small 24B 2507~\cite{liuVoxtral2025}; MiMo-V2 Omni and MiMo-V2.5~\cite{xiaomiMiMoModelRelease2026}; Nemotron 3 Nano Omni 30B A3B Reasoning~\cite{nvidiaNemotron3Nano2026}; Audio Flamingo Next and Audio Flamingo Next Think~\cite{ghoshAudioFlamingoNext2026};
Music Flamingo 2601 and Music Flamingo 2601 Think~\cite{ghoshMusicFlamingoScaling2025}. Throughout this paper, we use shortened model names. For example, Music Flamingo refers to Music Flamingo 2601.

Generation temperature is set to 0 where the API allows it. For Qwen3 Omni Thinking, we set the temperature to 0.6, Top P to 0.95, and Top K to 20 as recommended by the authors, since outputs at temperature 0 were very poor. MiMo-V2.5 does not support temperature control~\citep{XiaomiMiMoApi}.

\section{Results and Discussion}

\subsection{Screening}
\label{sec:screening}

Our experiments aim to resolve three uncertainties.

(i) Is parametric knowledge alone enough to test models' ability to recognise melodies at varying speeds, or do we need to provide examples in-context? Our experiments show that models are barely able to recognise melodies at normal speed, starting from the first note, using parametric knowledge alone.

(ii) Is there a preferred input modality? Should instructions be audio only or is a mix of text and interleaved audio files more effective? We find that text+audio prompting yields much higher accuracy scores than audio-only.

(iii) How many in-context examples are adequate for the tested \glspl{llm}? We ablate the in-context variant, testing eight, six and four in-context examples.

\paragraph{Zero shot} Results for both settings are in Tab.~\ref{tab:screening-zero-shot} in the Appendix. Most models in the text+audio setting classify only one out of eight melodies correctly, which is equal to closed-set chance. Qwen2 Audio and Audio Flamingo Think reach 25\% accuracy, and Qwen3.5 Omni Plus reaches 38\%.

In the audio-only setting, Gemini 3.1 Pro and Qwen3.5 Omni Flash reach 38\% accuracy, and Gemini 3 and 2.5 Flash reach 25\%. Despite music training, Music Flamingo and Audio Flamingo make at most one correct classification.

Overall, all models perform poorly and do not pass the screening phase, which requires models to classify $6/8$ melodies correctly.

\paragraph{In-context} In this setting, models were presented with melodies in context before being asked to classify a melody (multiple-choice). For the four- and six-reference variants, we test seven candidate sets because audio models are sensitive to option selection and order \citep{lopezRobustnessAssessmentLarge2025}.

\begin{figure}[!t]
  \centering
  \input{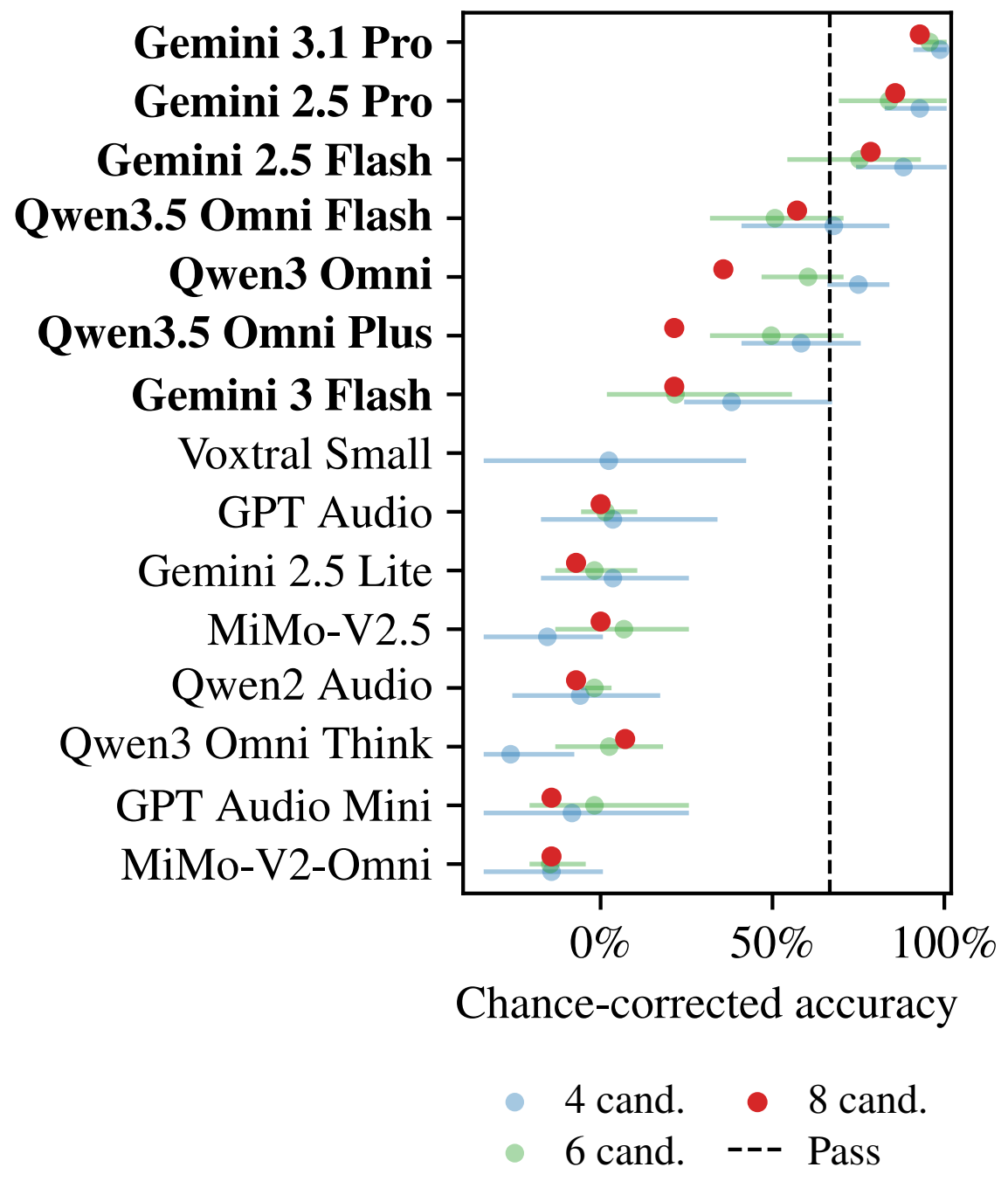}
  \caption{Chance-corrected text+audio screening accuracy with 4, 6, or 8
    candidate melodies at 320\,\textup{ms} per note (Sec.~\ref{sec:duration-probe}).
    For 4 and 6 candidates, points and whiskers show the mean and min--max across
    seven balanced candidate sets. The dashed line marks the pass criterion,
    chance-corrected for 4 candidates. Bold model names identify models that
    meet the screening criterion in at least one condition, using the mean
    or best candidate set.}
  \label{fig:screening-in-context-audio+text}
\end{figure}

Models perform much better in the text+audio setting than in the audio-only setting (Fig.~\ref{fig:screening-in-context-audio+text}). All models achieve higher overall scores and more models cross the selection threshold in some form. Both Gemini Pro models achieve accuracy close to 100\%, even with varying numbers of references, unlike in the audio-only setting. Gemini 2.5 Flash also crosses the selection threshold with different numbers of candidate melodies. Qwen3.5 Omni Flash and Plus, Qwen3 Omni, and Gemini 3 Flash crossed the screening threshold in some of the experiments, including their whiskers. These models are identified by bold names (Fig.~\ref{fig:screening-in-context-audio+text}). Most models perform equally well for all three candidate numbers. Noticeable differences can be observed for Qwen3 Omni and Qwen3.5 Omni Plus. Note that Voxtral Small only has valid runs for the setting with four in-context examples because it cannot process more than five audio files.

\begin{figure}[!t]
  \centering
  \input{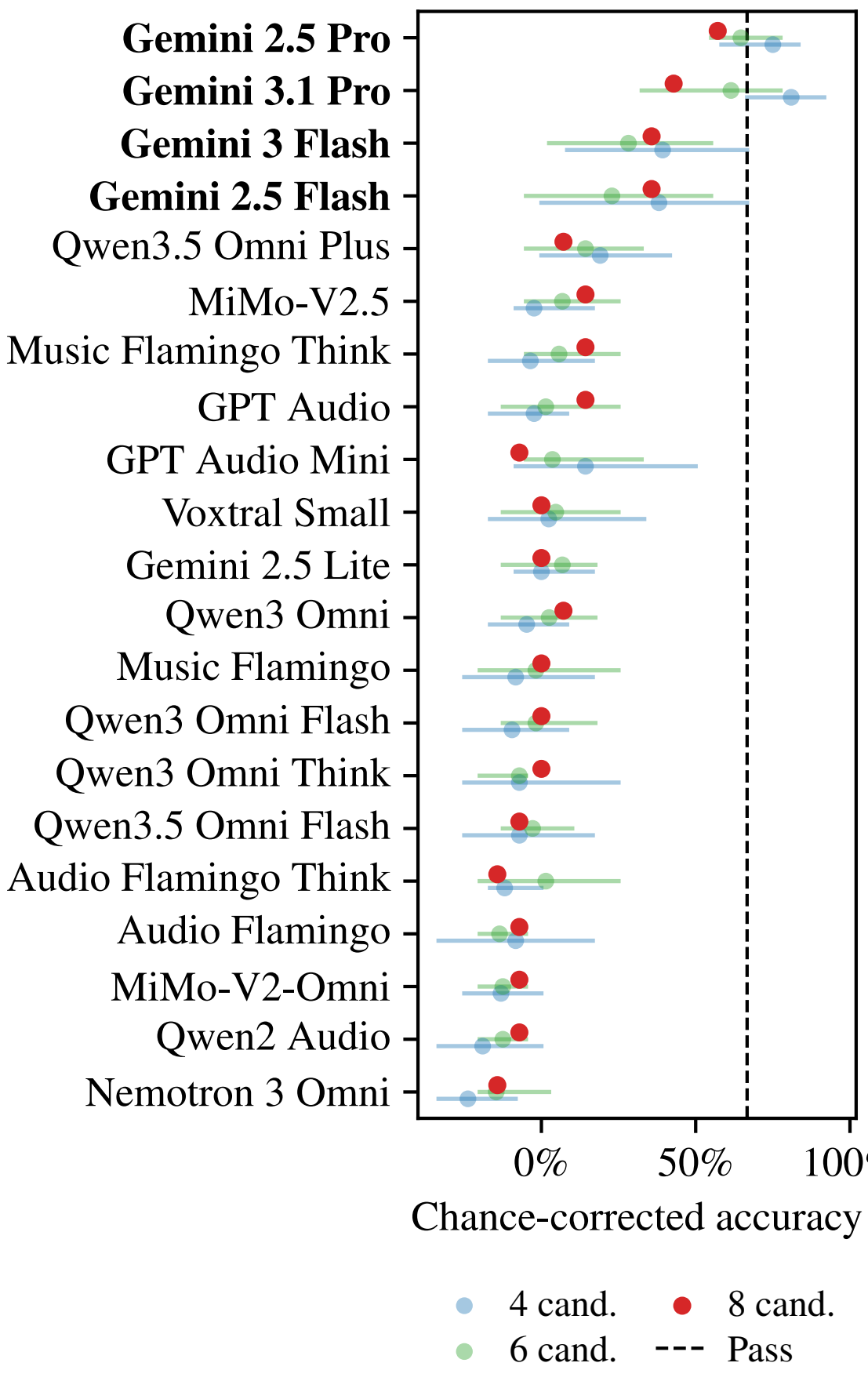}
  \caption{Chance-corrected audio-only screening accuracy under the design and
    encodings of Fig.~\ref{fig:screening-in-context-audio+text}. Bold model names
    identify models that meet the screening criterion in at least one condition,
    using the mean or best candidate set.}
  \label{fig:screening-in-context-audio-only}
\end{figure}

In contrast, models in the audio-only screening experiment perform poorly, with only four models crossing the acceptance threshold in some form (Fig.~\ref{fig:screening-in-context-audio-only}). While both Gemini Pro models clear the criterion on average with four candidate melodies, both Gemini Flash models reach it only with their best candidate sets (whiskers) in the four-reference ablation. We can observe that model performance is mostly comparable among the different numbers of in-context examples. The exceptions are the two Gemini Pro models, which, on average, perform better with fewer in-context examples --- even when comparing chance-corrected accuracy values. We find that Gemini Pro models deviate more often from the formatting instruction with eight reference melodies compared to six or four. We assume that this is driven by file length. A single experiment produces an almost three-minute audio file, which can dilute its opening instructions.

Therefore, the main experiments are conducted using in-context examples delivered as text+audio prompts. While the human study sought to identify a detection window, which requires strong melody recognition abilities, we are more interested in analysing how accuracy varies conditioned on the note duration of the target melody. We therefore pick all models that crossed the selection threshold, even if it was for a lower number of in-context examples or only the whiskers of the dot plots crossed the threshold (indicated by bold model names in Fig.~\ref{fig:screening-in-context-audio+text}). Still, to ensure comparability to the human study and given the overall similar results for different numbers of candidate melodies, we run the duration scaling experiments with eight in-context candidate melodies. Seven models pass screening and enter the duration scaling experiments: Gemini 3.1 Pro, Gemini 3 Flash, Gemini 2.5 Pro and Flash, Qwen3.5 Omni Plus and Flash, and Qwen3 Omni.

\subsection{Duration scaling}

\begin{figure}[!t]
  \centering
  \input{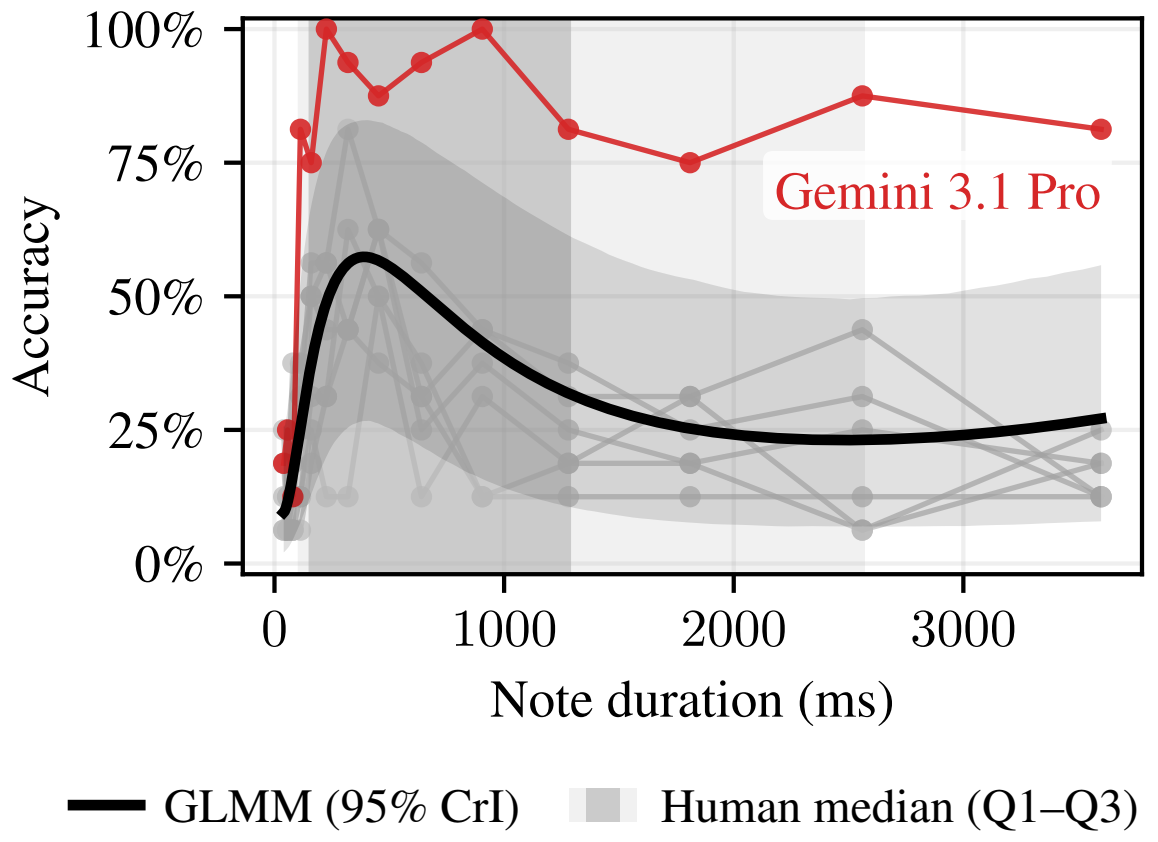}
  \caption{Accuracy versus target note duration for eight melodies
    ($n{=}1568$ trials). Targets begin at note 3 or 4 and are played twice. In-context references use 320\,\textup{ms} notes. Grey traces show the six other models that passed
    screening (Sec.~\ref{sec:screening}). Red highlights Gemini 3.1 Pro. The
    black curve and band show the
    population-level \gls{glmm} estimate and 95\% credible interval
    (CrI in the legend; Eq.~\ref{eq:duration-glmm}). Grey bands mark median
    (dark) and quartile (light) human limits (Sec.~\ref{sec:duration-probe}).}
  \label{fig:duration-accuracy}
\end{figure}

Surprisingly, most models, though with reduced overall performance, peak around the human median perception window (Fig.~\ref{fig:duration-accuracy}). We summarise model performance over all trials and models using the \gls{glmm} described in Eq.~\ref{eq:duration-glmm}. The broad posterior interval shown in the figure is driven by inter-model differences, most prominently by Gemini 3.1 Pro, which is the only model that maintains a high accuracy score regardless of note duration, though even this model shows a clear peak matching the median human recognition window.

All tested models except Gemini 3.1 Pro have difficulty confidently classifying a simple melody that varies in speed, though they perform much better when the in-context examples (played at 320\,\textup{ms}) and target melody are similar in speed. Performance for all models is poor if the target melody is played at very short note durations of less than 160\,\textup{ms}. A likely reason is that tokenisation is often performed in 160\,\textup{ms} chunks.

Gemini 3.1 Pro may be the only model with superhuman melody recognition. Even at very long note durations, where human perception shifts from a coherent melody to individual notes, it still achieves around 80\% accuracy.
However, the human interval is based on non-musicians and does not reflect the transcription-based strategy described in Sec.~\ref{sec:duration-probe}.
Such a strategy could explain Gemini 3.1 Pro's plateau, but without its reasoning traces, we cannot distinguish transcription from superhuman temporal perception.

\paragraph{Ablation of in-context note duration}

\begin{figure}[!t]
  \centering
  \input{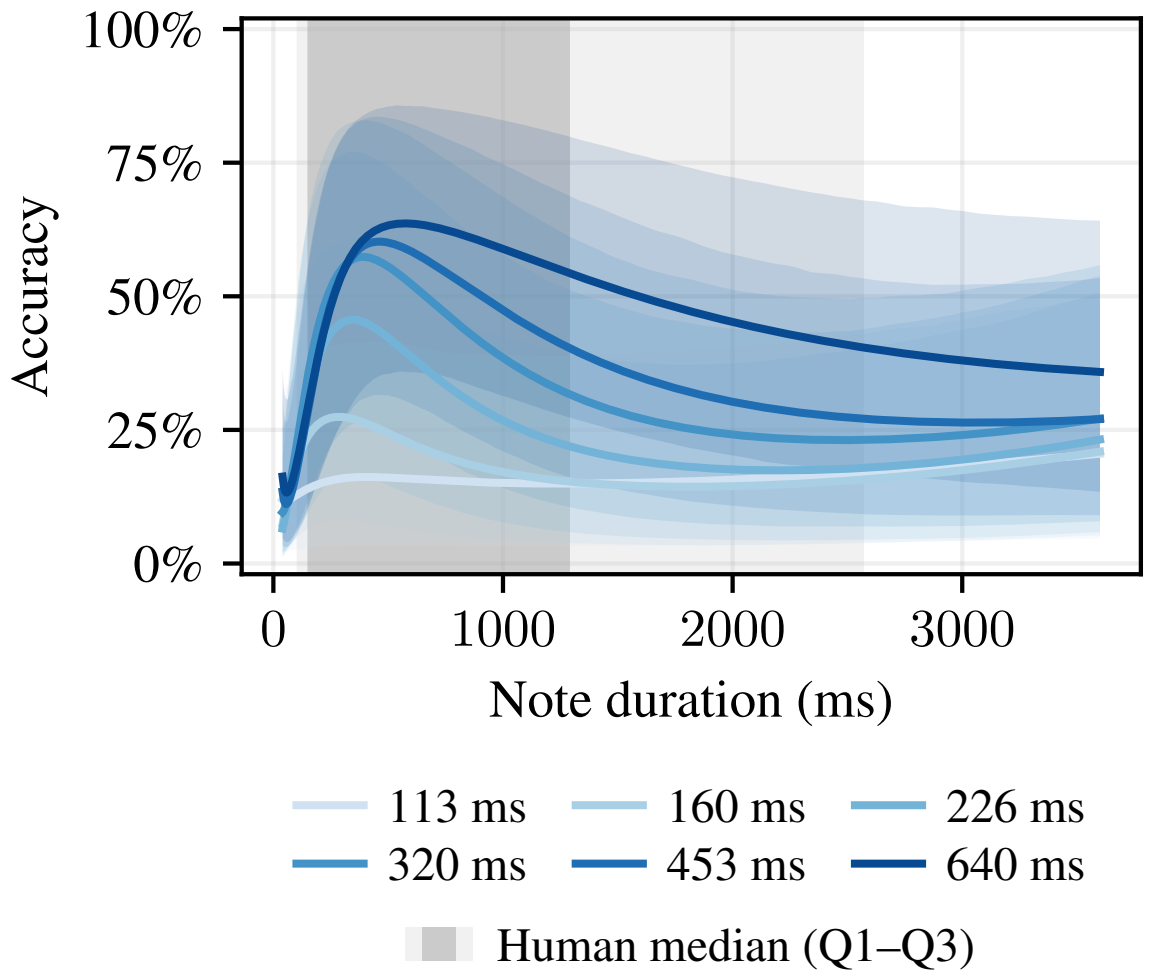}
  \caption{Population-level \gls{glmm} accuracy versus target note duration,
    one fit per in-context reference duration (113--640\,\textup{ms};
    colours). Curves and bands show posterior means and 95\% credible intervals.
    Human bands as in Fig.~\ref{fig:duration-accuracy}.}
  \label{fig:duration-accuracy-ablation}
\end{figure}

Accuracy seems to peak near 320\,\textup{ms} (Fig.~\ref{fig:duration-accuracy}). To answer if in-context note duration influences peak melody recognition, we test in-context note durations going from 113--640\,\textup{ms} \(\in D\), and refit the \gls{glmm} to assess how model performance behaves as a function of changed candidate melody speed. We can observe that overall model performance increases when we increase in-context melodies' note durations, and that the peak in accuracy shifts to the right (Fig.~\ref{fig:duration-accuracy-ablation}). On average, models are unable to make any correct prediction if in-context examples have a note duration of 113\,\textup{ms}, as for target melodies in the same speed in the main experiment.

\paragraph{Inter-melody differences}

\begin{figure}[!t]
  \centering
  \input{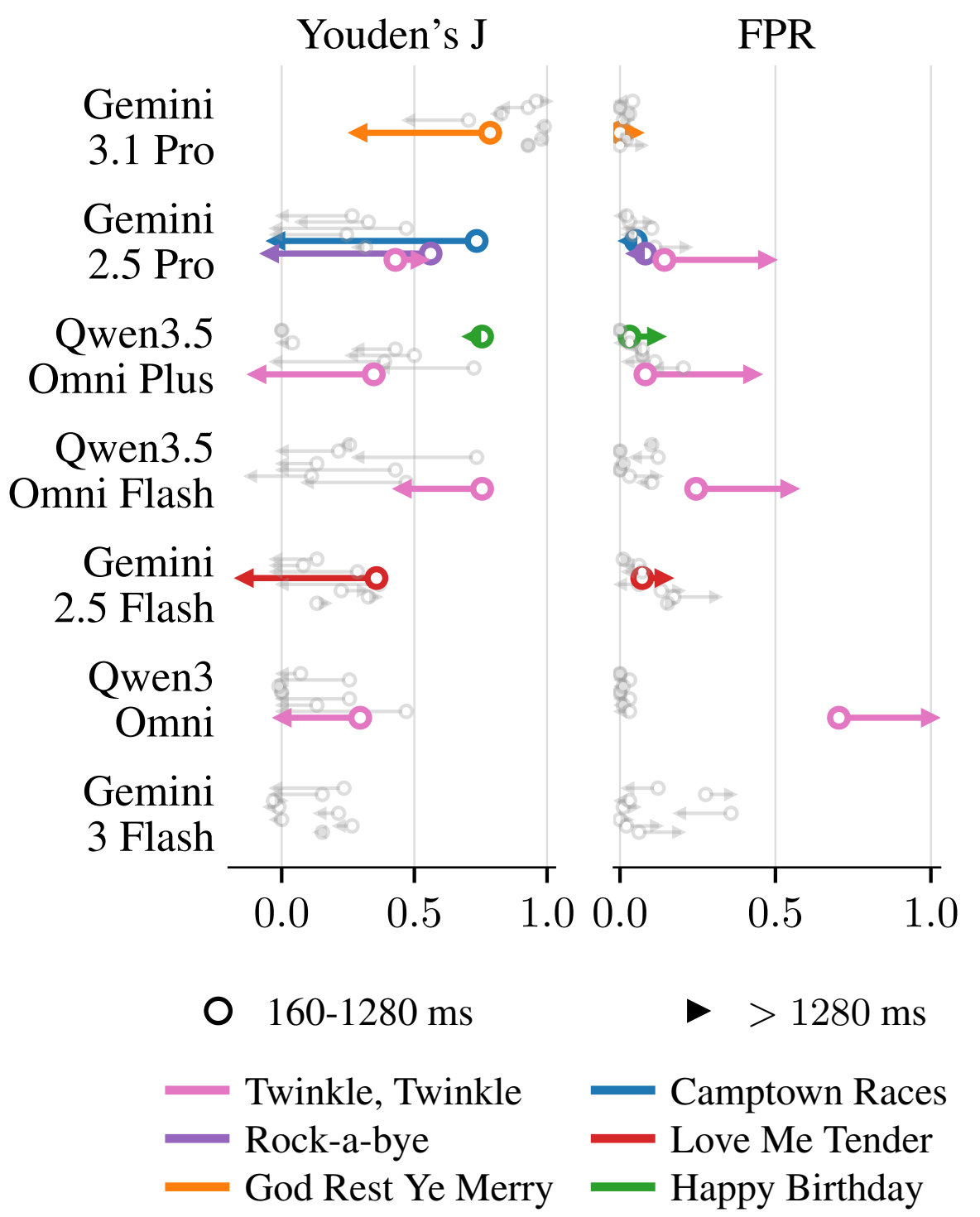}
  \caption{Per-model, per-melody changes in Youden's J and \gls{fpr} from
    target note durations of 160--1280\,\textup{ms} (circles) to
    ${>}1280$\,\textup{ms} (arrowheads). Coloured arrows highlight atypical
    melody--model pairs, such as increasing
    \gls{fpr} or declining Youden's J. Grey arrows show
    the remaining melody--model pairs.}
  \label{fig:duration-accuracy-bias}
\end{figure}

Our work primarily compares model performance to the median human recognition band. This band differs across melodies. The median band for Happy Birthday (160--1810\,\textup{ms}) is wider than that for Skip to My Lou (226--905\,\textup{ms}) for example. Several factors may contribute, such as prior familiarity with a melody and melody-specific salience. We therefore also investigate inter-melody differences in model performance in Fig.~\ref{fig:duration-accuracy-bias}. The figure shows Youden's J~\citep{youdenIndexRatingDiagnostic1950} and the \gls{fpr} for each model and melody for the human median recognition band (160--1280\,\textup{ms}) and above it (${>}1280$\,\textup{ms}). Youden's J is defined as \(J = \text{true positive rate} - \text{false positive rate}\). Assuming some melodies are easier and therefore over-predicted, Youden's J balances inflated true-positive rates against the \gls{fpr}.

In Fig.~\ref{fig:duration-accuracy-bias}, three phenomena emerge. (i) Twinkle, Twinkle, Little Star is over-predicted (high \gls{fpr}). Gemini 2.5 Pro, Qwen3.5 Omni Plus and Flash, and Qwen3 Omni show increased \gls{fpr} when the target melody is above the human median band. Qwen3 Omni already over-predicts it within the band. (ii) Some melodies are much harder to classify as note duration increases, but without an increase in \gls{fpr}. This occurs for Rock-a-bye Baby and Camptown Races in Gemini 2.5 Pro, and Love Me Tender in Gemini 3.1 Pro and Gemini 2.5 Flash. (iii) Happy Birthday is well recognised with no significant \gls{fpr} increase for Qwen3.5 Omni Plus. 

A possible explanation for the over-prediction of Twinkle, Twinkle, Little Star in Fig.~\ref{fig:duration-accuracy-bias} is that it is an artefact of the in-context examples. We therefore vary their note duration and compare median \gls{fpr} and median Youden's J for each of the six reference durations (113--640\,\textup{ms}) at target durations above the human median band (Table~\ref{tab:duration-bias-plateau-twinkle-by-ref}). Twinkle's \gls{fpr} is consistently higher, ranging from 0.36 to 0.60 compared to 0.03--0.05 for the other seven melodies. Its median Youden's J is likewise higher at every duration; the largest gap occurs at 320\,\textup{ms} (0.30 vs.\ 0.17).

There are various explanations why models would exhibit melody-specific biases as shown in Fig.~\ref{fig:duration-accuracy-bias}. Commonly, such biases can be traced back to the training data. Since the models' training data are not accessible, we perform a surrogate data analysis on YouTube. Although the number of uploads matters more for training than view counts do, we also report view counts, since they may inform data selection.

We obtained video upload rates via the YouTube API across four fixed 24-hour publication windows in 2023--2024 in March and September. We exclude the winter months, assuming that God Rest Ye Merry Gentlemen could be uploaded more often around Christmas. Searches included alternative titles for each melody. Duplicate videos were removed within melodies, and title-based matching excluded irrelevant search results. We report the number of matching videos, unique uploading channels, and median views in Tab.~\ref{tab:youtube-stats}.

\begin{table}[!t]
  \centering
  \small
\begin{tabular}{lrrr}
\toprule
Melody & Videos & Channels & \makecell{Median\\views} \\
\midrule
\makecell[l]{Twinkle Twinkle\\Little Star} & 122 & 103 & 4,105 \\
Love Me Tender & 47 & 42 & 95 \\
Happy Birthday & 36 & 34 & 56,925 \\
Yankee Doodle & 34 & 31 & 169 \\
Rock-a-bye Baby & 23 & 23 & 816 \\
Skip to My Lou & 15 & 15 & 105 \\
\makecell[l]{God Rest Ye\\Merry Gentlemen} & 11 & 10 & 56 \\
Camptown Races & 6 & 6 & 116 \\
\bottomrule
\end{tabular}

  \caption{YouTube search results for eight melodies, pooled across four fixed
    24-hour publication windows in 2023--2024: title-matched videos, unique
    channels, and median views per matched video. Aliases are pooled and
    duplicate videos removed.}
  \label{tab:youtube-stats}
\end{table}

Twinkle, Twinkle, Little Star has the most title matches (122) and unique uploading channels (103). With 4,105 views, it is the second-most-listened-to song in the analysis. The many uploading channels confirm genuine interest in the melody and argue against a few channels inflating upload counts. Happy Birthday ranks first in views (56,925) but has fewer title matches. Twinkle's high title-match count may reflect that the melody appears in songs under several aliases. If the model training data resemble YouTube and our query is representative, models likely encountered the melody more often than others, potentially explaining its higher \gls{fpr} in Fig.~\ref{fig:duration-accuracy-bias}. Because the melody has different names, deduplication may not have removed many training instances. God Rest Ye Merry Gentlemen and Camptown Races also became harder to classify as note duration increased; our YouTube query finds the fewest videos for these melodies (eleven and six, respectively).

\begin{figure}[t]
  \centering
  \input{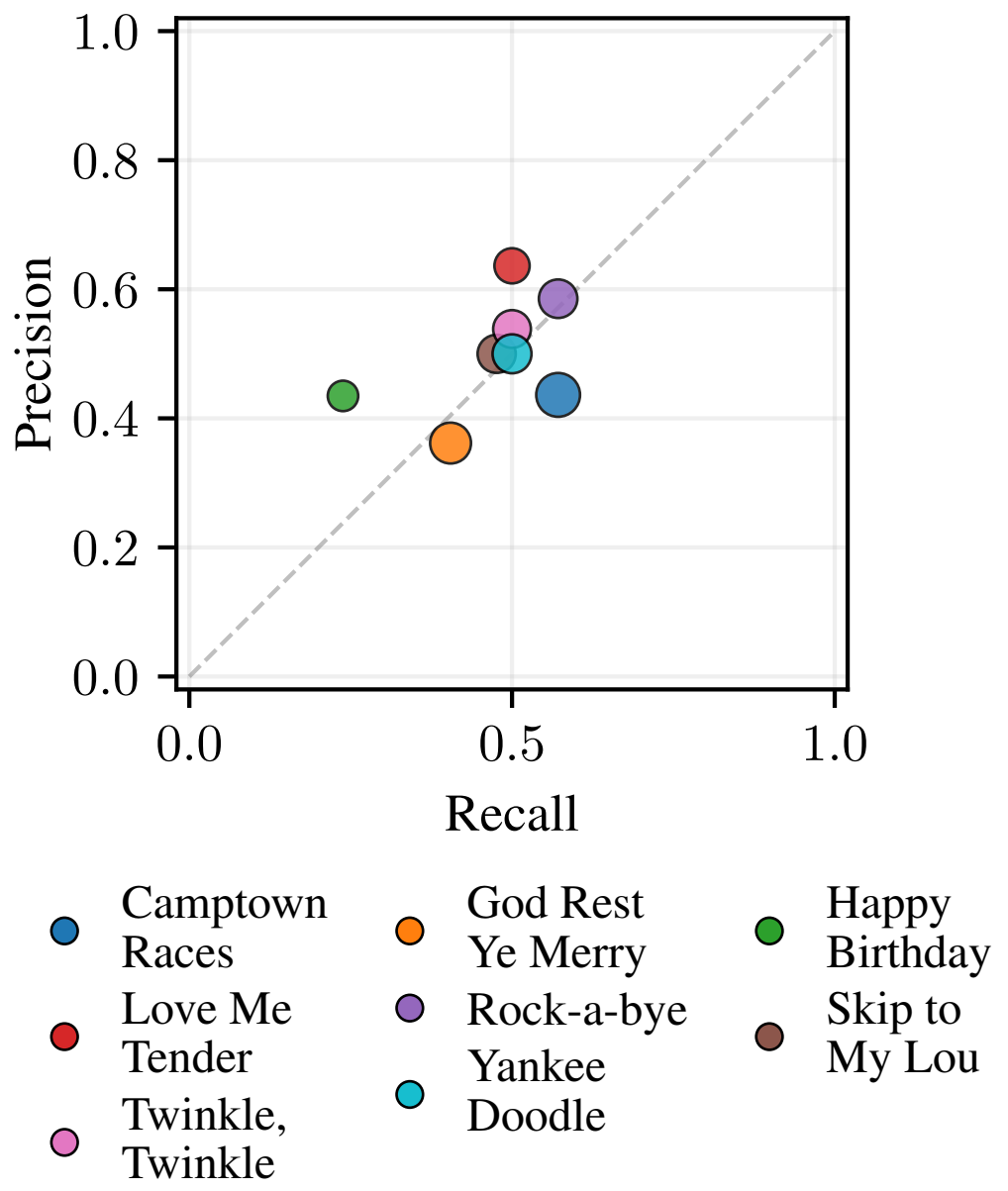}
  \caption{Per-melody precision and recall in the neutral-code ablation,
    pooled over the trials of all screened models at target note durations
    ${>}1280$\,\textup{ms}. Marker area scales with how often the title was
    predicted. The dashed diagonal denotes equal precision and recall.}
  \label{fig:precision-recall-random-titles}
\end{figure}

We explore whether the melody bias can be mitigated by replacing music titles with neutral letter--digit codes, such as ``D4''. Indeed, over-prediction of melodies decreases and overall performance increases. Fig.~\ref{fig:precision-recall-random-titles} shows that no melody has an overly high recall or precision. In comparison, prompting models with the full melody name leads to an over-prediction of Twinkle, Twinkle, Little Star for some models. The model performance is also better when using codes instead of melody titles.

\setcounter{topnumber}{1}

\begin{figure}[t]
  \centering
  \input{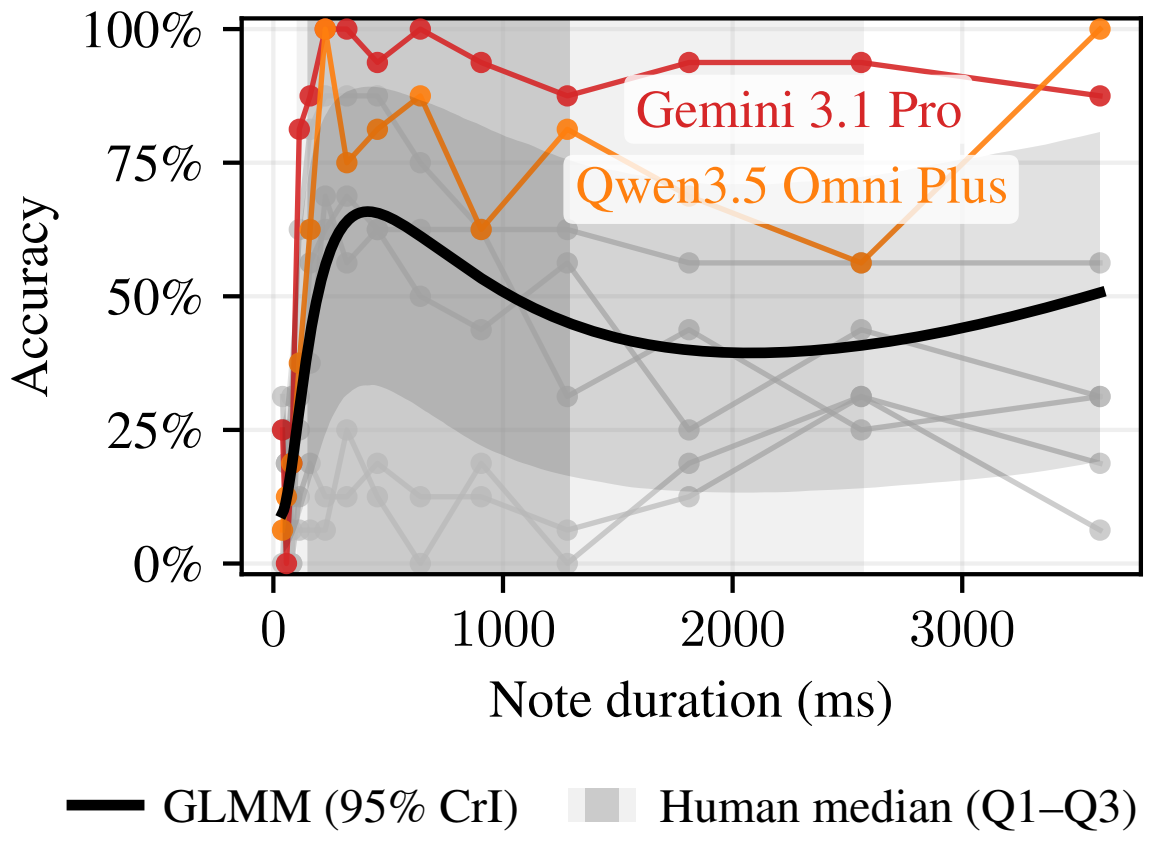}
  \caption{Accuracy versus target note duration when melody titles are replaced
    with neutral letter--digit codes. The setup, model selection, and encodings
    follow Fig.~\ref{fig:duration-accuracy}. Red and orange highlight Gemini
    3.1 Pro and Qwen3.5 Omni Plus, respectively.}
  \label{fig:duration-accuracy-random-titles}
\end{figure}

The \gls{glmm} fitted curve peaks above 60\% (Fig.~\ref{fig:duration-accuracy-random-titles}), versus around 50\% with full titles (Fig.~\ref{fig:duration-accuracy}). Gemini 3.1 Pro and Qwen3.5 Omni Plus lie above the fit's credible interval.

\paragraph{More melodies}

While eight melodies may yield robust results about human time perception, inter-melody bias findings as observed in Fig.~\ref{fig:duration-accuracy-bias} cast doubt on the robustness of our results. Therefore, we add eleven more songs that are also considered to be widely known (in the English-speaking world), ensuring that melody recognition is only constrained by note duration, not by familiarity with the melody. These include America, Frère Jacques, Here Comes the Bride, Jingle Bells, Jolly Good Fellow, London Bridge, Mary Had a Little Lamb, Old MacDonald, Row, Row, Row Your Boat, The Star-Spangled Banner, and Three Blind Mice \citep{gfellerRecognitionFamiliarMelodies2002, kangDevelopmentValidationUniversity2009}.

Extending the set beyond the eight melodies of \citet{warrenMelodicNonmelodicSequences1991} reproduces the same pattern (Appendix, Fig.~\ref{fig:duration-accuracy-extended}): all models peak in the same region with similar accuracy, while Gemini 3.1 Pro shows a similar but more stable peak and plateau.

\paragraph{Scientific pitch notation}

\begin{figure}[t]
  \centering
  \input{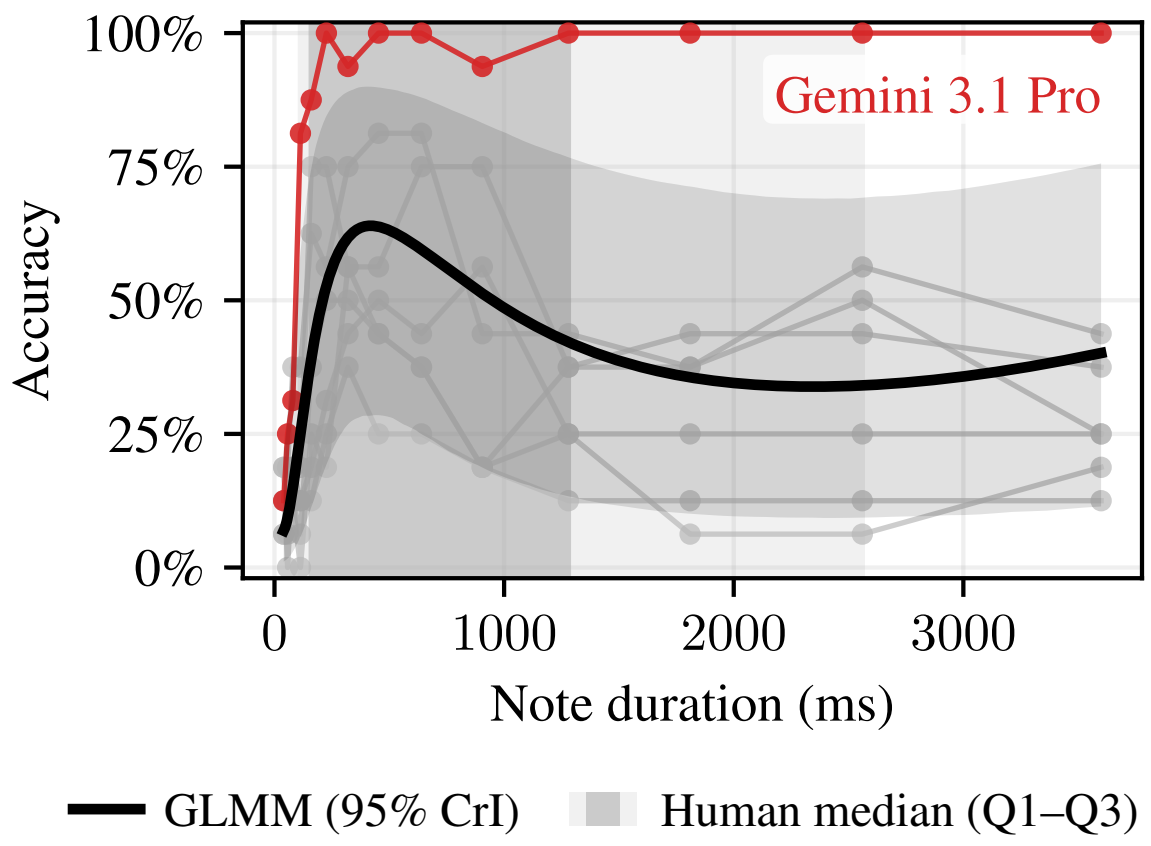}
  \caption{Accuracy versus target note duration with scientific-pitch notation
    for each in-context reference. The setup, model selection, and
    encodings follow Fig.~\ref{fig:duration-accuracy}.}
  \label{fig:duration-accuracy-text-notes}
\end{figure}

We next report the transcription ablation introduced in Sec.~\ref{sec:duration-probe}. When we supply scientific pitch notation for each melody, average peak accuracy rises from just above 50\% to around 60\% (Figs.~\ref{fig:duration-accuracy} and~\ref{fig:duration-accuracy-text-notes}). Gemini 3.1 Pro remains near-perfect even at 3600\,\textup{ms} per note. The other models improve but still decline with duration. Thus, notation helps most models, but only Gemini 3.1 Pro sustains high long-duration accuracy, as human musicians do; whether it transcribes internally remains unclear.

\section{Conclusion}

In this work, we test the limits of duration perception in \glspl{mllm} using a melody-recognition task. 
Most of the 21 models tested struggle to recognise familiar melodies, even with labelled in-context examples. 
The few that succeed peak within the human recognition band. However, this resemblance does not necessarily reflect a comparable perceptual limit, as the recognition window shifts with the speed of the references. 
Recognition also depends on what a melody is called: replacing familiar titles with neutral codes reduces prediction bias and improves performance. 
Even with scientific pitch notation, most models cannot sustain recognition at long note durations as human musicians do. 
Gemini 3.1 Pro is an exception, recognising melodies even at 3600 ms per note, though whether this reflects greater perceptual capacity or reasoning strategies remains unclear. 
\Glspl{mllm} have the potential to surpass human perceptual limits, and this work begins to stake out these boundaries.

\section{Limitations}

Human participants heard melodies in interactive loops and relied on prior familiarity, whereas models received fixed trials and in-context references because their zero-shot recognition was insufficient.
The resulting curves should therefore be interpreted as behavioural patterns rather than as directly equivalent perceptual thresholds.
Once interactive models become the norm, we should test them more similarly to the human setup.
Furthermore, our stimuli are isochronous sine-tone versions of familiar melodies drawn largely from English-speaking musical culture, so the findings may not generalise to natural recordings, unfamiliar or non-Western music, or duration perception beyond melodic grouping.
Finally, black-box behaviour cannot reveal the mechanisms underlying model performance.
Sustained accuracy may arise from temporal grouping, pitch tracking, transcription, or retrieval from training data; the lack of access to these data prevents us from distinguishing among these explanations.

\section*{Acknowledgements}

Auss Abbood was supported by the Economic and Social Research Council [ES/Y001788/1].
This work was supported by the Google Cloud Research Credits program.

\bibliographystyle{acl_natbib}
\bibliography{references-fixed}

\clearpage
\appendix
\section{Experiment setup}

We used GNU Parallel~\citep{tangeGNUParallel202604222026} to run experiment sweeps across models and experimental conditions.

\subsection{Melodies}
\label{app:melodies}

Notes in melodies were rendered as sine waves at a sampling rate of 44.1\,\textup{kHz}, with a peak amplitude of 0.2, 5\,\textup{ms} linear attack and release ramps to avoid onset/offset clicks, and a 20\,\textup{ms} silent gap after each sounding portion. Note durations are specified onset-to-onset (i.e., each interval includes both the tone and the following gap). The resulting waveforms were saved as mono WAV files.

\begin{table}[t]
  \centering
  \small
\setlength{\tabcolsep}{4pt}
\begin{tabular}{@{}>{\raggedright\arraybackslash}p{0.34\linewidth}>{\raggedright\arraybackslash}p{\dimexpr\linewidth-0.34\linewidth-2\tabcolsep\relax}@{}}
\toprule
Melody & Notes \\
\midrule
Twinkle Twinkle Little Star & F\#4 F\#4 C\#5 C\#5 D\#5 D\#5 C\#5 C\#5 \\
Happy Birthday & G\#4 G\#4 A\#4 G\#4 C\#5 C5 \\
God Rest Ye Merry Gentlemen & G4 G4 D5 D5 C5 Bb4 A4 G4 \\
Rock-a-bye Baby & E4 G4 E5 D5 C5 E4 G4 C5 B4 \\
Camptown Races & C\#5 C\#5 A\#4 C\#5 D\#5 C\#5 A\#4 \\
Yankee Doodle & Bb4 Bb4 C5 D5 Bb4 D5 C5 \\
Love Me Tender & F4 Bb4 A4 Bb4 C5 G4 C5 \\
Skip to My Lou & B4 B4 G4 G4 B4 B4 D5 D5 \\
\bottomrule
\end{tabular}

  \caption{The original eight melodies and their scientific-pitch sequences,
    from \citet{warrenMelodicNonmelodicSequences1991}.}
  \label{tab:melodies-original}
\end{table}

\begin{table}[t]
  \centering
  \small
\setlength{\tabcolsep}{4pt}
\begin{tabular}{@{}>{\raggedright\arraybackslash}p{0.34\linewidth}>{\raggedright\arraybackslash}p{\dimexpr\linewidth-0.34\linewidth-2\tabcolsep\relax}@{}}
\toprule
Melody & Notes \\
\midrule
America & G4 G4 A4 F\#4 G4 A4 \\
Frere Jacques & F4 G4 A4 F4 F4 G4 A4 F4 \\
Here Comes the Bride & F4 Bb4 Bb4 Bb4 F4 C5 A4 Bb4 \\
Jingle Bells & A4 A4 A4 C5 F4 G4 A4 \\
Jolly Good Fellow & A4 A4 A4 G4 A4 Bb4 A4 \\
London Bridge & A4 B4 A4 G4 F\#4 G4 A4 \\
Mary Had a Little Lamb & B4 A4 G4 A4 B4 B4 B4 \\
Old Mac Donald & A4 A4 A4 E4 F\#4 F\#4 E4 \\
Row Row Row & D4 D4 D4 E4 F\#4 F\#4 E4 F\#4 \\
Star-Spangled Banner & A4 F\#4 D4 F\#4 A4 D5 \\
Three Blind Mice & F\#4 E4 D4 F\#4 E4 D4 \\
\bottomrule
\end{tabular}

  \caption{Eleven widely familiar melodies added to the original set, with
    their scientific-pitch sequences. Selection is based on
    \citet{gfellerRecognitionFamiliarMelodies2002} and
    \citet{kangDevelopmentValidationUniversity2009}.}
  \label{tab:melodies-extended}
\end{table}

\subsection{Prompts}
\label{app:prompts}

This section contains the prompts used for the screening phase, which include audio-only and text+audio formatted prompts in a zero-shot setting (Figs.~\ref{fig:prompt-zero-shot-text-audio} \& \ref{fig:prompt-zero-shot-audio-only}) and an in-context setting (Figs.~\ref{fig:prompt-few-shot-text-audio} \& \ref{fig:prompt-few-shot-audio-only}). NVIDIA's Audio Flamingo models were prompted differently in the audio-only setting. According to their model card, the model should not be prompted with only an audio file. Therefore, we prompt them with the instruction \texttt{You will receive one audio file for a melody recognition task. The file contains spoken instructions, optional labelled reference melodies, and then a target melody. Listen to the full audio. Classify only the final target melody. Answer with exactly one title from the reference titles mentioned.} before supplying the WAV file.

\begin{figure}[h!]
  \centering
  \input{figures/prompt_zero_shot_text_audio}
  \caption{Zero-shot text+audio prompt used for screening. Text instructions
    precede the target audio. Candidate-title order is randomised per trial.}
  \label{fig:prompt-zero-shot-text-audio}
\end{figure}

\begin{figure}[h!]
  \centering
  \input{figures/prompt_zero_shot_audio_only}
  \caption{Zero-shot audio-only prompt used for screening. Text-to-speech
    instructions and target audio are concatenated into one WAV\@. Differences
    from Fig.~\ref{fig:prompt-zero-shot-text-audio} are bold. Candidate-title
    order is randomised per trial.}
  \label{fig:prompt-zero-shot-audio-only}
\end{figure}

\begin{figure}[h!]
  \centering
  \input{figures/prompt_few_shot_text_audio}
  \caption{In-context text+audio prompt used for screening and duration-grid
    experiments. The displayed reference list is abbreviated with [\dots].
    Experiments use eight references unless noted. The prompt remains fixed
    while the target clip's melody, entry point, and/or note duration varies.}
  \label{fig:prompt-few-shot-text-audio}
\end{figure}

\begin{figure}[h!]
  \centering
  \input{figures/prompt_few_shot_audio_only}
  \caption{In-context audio-only prompt used for screening. Text-to-speech
    instructions, labelled reference clips, and target audio are concatenated
    into one WAV\@. Differences from Fig.~\ref{fig:prompt-few-shot-text-audio}
    are bold. Experiments use eight references unless noted.}
  \label{fig:prompt-few-shot-audio-only}
\end{figure}

Spoken instructions use OpenAI's \texttt{gpt-4o-mini-tts} (voice \texttt{alloy}). They are concatenated with the labelled reference clips and the target melody, with 500\,ms of silence between consecutive segments.

The main grid experiments reuse the in-context text+audio prompt (Fig.~\ref{fig:prompt-few-shot-text-audio}) with different note durations for the target melody. The ablation that replaces melody titles with neutral letter--digit codes is identical to Fig.~\ref{fig:prompt-few-shot-text-audio} except for the melody titles. The ablation prompt that provides the scientific pitch notation (Fig.~\ref{fig:prompt-few-shot-text-notes}) extends the main prompt (Fig.~\ref{fig:prompt-few-shot-text-audio}). Differences are highlighted in the Figure.

\begin{figure}[h!]
  \centering
  \input{figures/prompt_few_shot_text_notes}
  \caption{In-context text+audio prompt with scientific-pitch sequences appended
    to each reference. Text shared with
    Fig.~\ref{fig:prompt-few-shot-text-audio} is elided; additions are bold.}
  \label{fig:prompt-few-shot-text-notes}
\end{figure}

\subsection{Scoring}
\label{app:scoring}
A trial is correct only if the response names exactly one melody from the closed set for that trial. Matching is case-insensitive, ignores punctuation, and uses word boundaries, so a title must appear as a whole phrase. We also accept pre-specified aliases: Twinkle, Twinkle, Little Star (ABC Song, Alphabet Song, Baa Baa Black Sheep), Happy Birthday (Happy Birthday To You), God Rest Ye Merry Gentlemen (God Rest You Merry Gentlemen, with or without a comma), and Love Me Tender (Aura Lee, Aura Lea). All other tunes are scored only on their canonical title. For thinking models, we discard <think>…</think> and score only the remainder. An unclosed think block, no recognised title, or two or more distinct titles count as incorrect. In the neutral-code ablation, the same rule applies to the letter-digit codes, not the real names.

\subsection{GLMM}
\label{app:glmm}

We fit the duration \gls{glmm} in Eq.~\eqref{eq:duration-glmm} with four NUTS chains (1000 warmup iterations and 1000 draws, \texttt{target\_accept}${=}0.95$, seed 0) and compare cubic $B$-splines with $K \in \{3,4,5,6,7\}$ by PSIS-LOO on the main eight-melody grid ($n{=}1568$ trials). $K{=}5$ has the highest ELPD\@; $K{=}4$ is within one standard error of that fit ($\Delta\mathrm{ELPD}{=}2.7$, SE of the difference ${=}3.2$) and is the smallest such $K$, so all reported curves use $K{=}4$. The shaded band is the 95\% posterior interval of population-level accuracy after integrating out the model and melody intercepts.

\newpage
\section{Further results}
\label{app:results}

\subsection{Extended melody list}

\begin{figure}[H]
  \centering
  \input{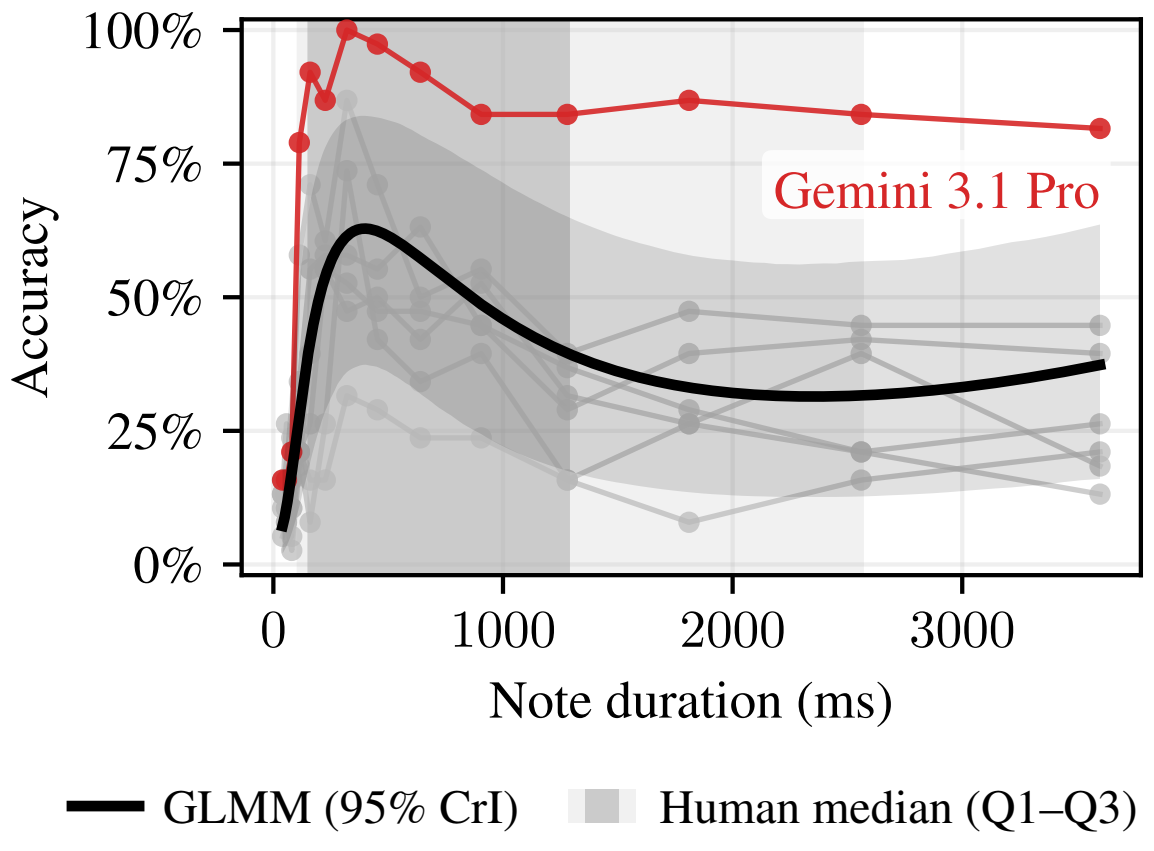}
  \caption{Accuracy versus target note duration for the extended set of 19
    melodies. The setup, model selection, and encodings follow
    Fig.~\ref{fig:duration-accuracy}.}
  \label{fig:duration-accuracy-extended}
\end{figure}

\subsection{Zero-shot screening}

\begin{table}[H]
  \centering
  \small
\begin{tabular}{lrr}
\toprule
 & \multicolumn{2}{c}{Acc. (\%)} \\
\cmidrule(lr){2-3}
Model & Text+audio & Audio-only \\
\midrule
Qwen3.5 Omni Plus & 38\% & 25\% \\
Gemini 3.1 Pro & 12\% & 38\% \\
Qwen3.5 Omni Flash & 12\% & 38\% \\
Gemini 2.5 Flash & 12\% & 25\% \\
Gemini 3 Flash & 12\% & 25\% \\
Audio Flamingo Think & 25\% & 0\% \\
Gemini 2.5 Pro & 12\% & 12\% \\
GPT Audio Mini & 12\% & 12\% \\
Qwen2 Audio & 25\% & 0\% \\
Qwen3 Omni & 12\% & 12\% \\
Qwen3 Omni Think & 12\% & 12\% \\
Audio Flamingo & 0\% & 12\% \\
GPT Audio & 12\% & 0\% \\
MiMo-V2-Omni & 12\% & 0\% \\
MiMo-V2.5 & 12\% & 0\% \\
Music Flamingo & 12\% & 0\% \\
Music Flamingo Think & 12\% & 0\% \\
Nemotron 3 Omni & 0\% & 12\% \\
Qwen3 Omni Flash & 0\% & 12\% \\
Gemini 2.5 Lite & 0\% & 0\% \\
Voxtral Small & 0\% & 0\% \\
\bottomrule
\end{tabular}

  \caption{Zero-shot screening accuracy over eight melody trials per modality
    (eight-way closed set; chance ${=}12.5\%$). Targets begin at the first note.
    \textbf{Text+audio} uses text instructions followed by target audio.
    \textbf{Audio-only} concatenates text-to-speech instructions and target
    audio into one WAV.}
  \label{tab:screening-zero-shot}
\end{table}

\subsection{Over-prediction of Twinkle, Twinkle, Little Star}

\begin{table}[H]
  \centering
  \small
\begin{tabular}{lrrrr}
\toprule
 & \multicolumn{2}{c}{FPR} & \multicolumn{2}{c}{Youden's J} \\
\cmidrule(lr){2-3}\cmidrule(lr){4-5}
Ref.\ (ms) & Twinkle & Others & Twinkle & Others \\
\midrule
113 & 0.60 & 0.03 & 0.17 & 0.07 \\
160 & 0.53 & 0.04 & 0.18 & 0.06 \\
226 & 0.48 & 0.05 & 0.24 & 0.15 \\
320 & 0.41 & 0.04 & 0.30 & 0.17 \\
453 & 0.41 & 0.05 & 0.21 & 0.19 \\
640 & 0.36 & 0.04 & 0.33 & 0.27 \\
\bottomrule
\end{tabular}

  \caption{Median \gls{fpr} and median Youden's J at target note durations
    ${>}1280$\,\textup{ms}, pooled over the trials of all screened models,
    comparing Twinkle, Twinkle, Little Star with the median of the remaining
    seven melodies, by in-context reference note duration. Higher \gls{fpr}
    indicates stronger over-prediction; higher Youden's J indicates better
    discrimination.}
  \label{tab:duration-bias-plateau-twinkle-by-ref}
\end{table}

\end{document}